%% file: main.tex
\documentclass[11pt,a4paper]{article}
\usepackage[hyperref]{naaclhlt2018}
\usepackage{times}
\usepackage{latexsym}

\usepackage{url}

\usepackage{amsmath}
\usepackage{multirow}

\usepackage{graphicx}
\usepackage{caption}
\usepackage{subcaption}

\graphicspath{{./figs/}}

\usepackage{tabularx}

\usepackage{textcomp}

\aclfinalcopy 

\title{Implicit Argument Prediction with Event Knowledge}

\author{Pengxiang Cheng \\
  Department of Computer Science \\
  The University of Texas at Austin \\
  {\tt pxcheng@cs.utexas.edu} \\\And
  Katrin Erk \\
  Department of Linguistics \\
  The University of Texas at Austin \\
  {\tt katrin.erk@mail.utexas.edu} \\}

\date{}

\begin{document}
\maketitle
\begin{abstract}
  \input{0-abstract.tex}
\end{abstract}

\section{Introduction}
\label{sec::intro}
\input{1-intro.tex}

\section{Related Work}
\label{sec::related}
\input{2-related.tex}

\section{The Argument Cloze Task}
\label{sec::task}
\input{3-task.tex}

\section{Methods}
\label{sec::methods}
\input{4-methods.tex}

\section{Evaluation Datasets}
\label{sec::datasets}
\input{5-datasets.tex}

\section{Experiments}
\label{sec::exp}
\input{6-experiments.tex}

\section{Conclusion}
\label{sec::conclusion}
\input{7-conclusion.tex}

\section*{Acknowledgments}
This research was supported by NSF grant IIS 1523637. We also acknowledge the Texas Advanced Computing Center for providing grid resources that contributed to these results, and we would like to thank the anonymous reviewers for their valuable feedback.

\bibliography{ref}
\bibliographystyle{acl_natbib}

\appendix

\clearpage

\input{supplemental.tex}

\end{document}

%% file: 0-abstract.tex
Implicit arguments are not syntactically connected to their
predicates, and are therefore hard to extract. Previous work has used
models with large numbers of features, evaluated on very small
datasets. We propose to train models for implicit argument prediction
on a simple cloze task, for which data can be generated automatically
at scale. This allows us to use a neural model, which draws on
narrative coherence and entity salience for predictions. We show
that our model has superior performance on both synthetic and natural
data. \footnote{Our code is available at \url{https://github.com/pxch/event_imp_arg}.}

%% file: 1-intro.tex
When parts of an event description in a text are missing, this event
cannot be easily extracted, and it cannot easily be found as the answer
to a question. This is the case with \emph{implicit arguments}, as in
this example from the reading comprehension dataset of \citet{hermann2015teaching}:
\begin{quote}
\textbf{Text:}  More than 2,600 people have been infected by Ebola in Liberia,
  Guinea, Sierra Leone and Nigeria since the \textit{outbreak} began in
  December, according to the World Health Organization. Nearly 1,500
  have \textit{died}.

\textbf{Question:} The X outbreak has killed nearly 1,500.
\end{quote}

In this example, it is Ebola that broke out, and Ebola was also the
cause of nearly 1,500 people dying, but the text does not state this
explicitly. \textit{Ebola} is an implicit argument of both \textit{outbreak} and \textit{die},
which is crucial to answering the question.  

We are particularly interested in implicit arguments that, like
\textit{Ebola} in this case, do appear in the text, but not as
syntactic arguments of their predicates. 
Event knowledge is key to determining implicit arguments. In our
example, diseases are maybe the single most typical things to \textit{break out}, 
and diseases also typically kill people. 

The task of identifying implicit arguments was first addressed by
\citet{Gerber2010ACL} and \citet{SemEval2010Task10}. However, the datasets for
the task were very small, and to our knowledge there has been very little
further development on the task since then. 

In this paper, we address
the data issue by training models for implicit argument prediction
on a simple cloze task, similar to the narrative cloze task \cite{Chambers2008ACL}, for which data can be generated automatically
at scale. This allows us to
train a neural network to perform the task, building on two
insights. First, event knowledge is crucial for implicit argument
detection. Therefore we build on models for narrative event
prediction~\cite{Granroth2016AAAI,Pichotta2016AAAI},
using them to judge how coherent the narrative would be when we fill
in a particular entity as the missing (implicit) argument. Second, the omitted arguments tend to be salient, as
\textit{Ebola} is in the text from which the above example is
taken. So in addition to narrative coherence, our model takes into
account entity salience~\cite{Dunietz2014EACL}. 

In an evaluation on a large automatically generated dataset, 
our model clearly outperforms even strong baselines, and we find 
salience features to be important to the success of the model. We also
evaluate against a variant of the \citet{Gerber2012CL} model that does not
rely on gold features, finding that our simple neural model
outperforms their much more complex model.

Our paper thus makes two major contributions. 1) We propose an argument cloze task to generate synthetic training data at scale for implicit argument prediction. 2) We show that neural event models for narrative schema prediction can be used on implicit argument prediction, and that a straightforward combination of event knowledge and entity salience can do well on the task.

%% file: 2-related.tex
While dependency parsing and semantic role labeling only deal with arguments that are available in the syntactic context of the predicate, implicit argument labeling seeks to find argument that are not syntactically connected to their predicates, like \textit{Ebola} in our introductory example.

The most relevant work on implicit argument prediction came from \citet{Gerber2010ACL}, who built an implicit arguments dataset by selecting 10 nominal predicates from NomBank \cite{meyers2004nombank} and manually annotating implicit arguments for all occurrences of these predicates. In an analysis of their data they found implicit arguments to be very frequent, as their annotation added 65\% more arguments to NomBank. \citet{Gerber2012CL} also trained a linear classifier for the task relying on many hand-crafted features, including  gold features from FrameNet \cite{baker1998framenet}, PropBank \cite{palmer2005propbank} and NomBank. This classifier has, to the best of our knowledge, not been outperformed by follow-up work \cite{Laparra2013ACL,Schenk2016NAACL,Do2017IJCNLP}. We evaluate on the Gerber and Chai dataset below. \citet{SemEval2010Task10} also introduced an implicit argument dataset, but we do not evaluate on it as it is even smaller and much more complex than \citet{Gerber2010ACL}. More recently, \citet{Modi2017TACL} introduced the referent cloze task, in which they predicted a manually removed discourse referent from a human annotated narrative text. This task is closely related to our argument cloze task.

Since we intend to exploit event knowledge in predicting implicit arguments, we here refer to recent work on statistical script learning, started by \citet{Chambers2008ACL,Chambers2009ACL}. They introduced the idea of using statistical information on coreference chains to induce prototypical sequences of narrative events and participants, which is related to the classical notion of a script \cite{schank1977scripts}. They also proposed the narrative cloze evaluation, in which one event is removed at random from a sequence of narrative events, then the missing event is predicted given all context events. We use a similar trick to define a cloze task for implicit argument prediction, discussed in Section \ref{sec::task}.

Many follow-up papers on script learning have used neural networks. \citet{Rudinger2015EMNLP} showed that sequences of events can be efficiently modeled by a log-bilinear language model. \citet{Pichotta2016AAAI,Pichotta2016ACL} used an LSTM to model a sequence of events. \citet{Granroth2016AAAI} built a network that produces an event representation by composing its components. To do the cloze task, they select the most probable event based on pairwise event coherence scores. For our task we want to do something similar: We want to predict how coherent a narrative would be with a particular entity candidate filling the implicit argument position. So we take the model of \citet{Granroth2016AAAI} as our starting point.

The \citet{hermann2015teaching} reading comprehension task, like our cloze task, requires systems to guess a removed entity. However in their case the entity is removed in a summary, not in the main text. In their case, the task typically amounts to finding a main text passage that paraphrases the sentence with the removed entity; this is not the case in our cloze task.



%% file: 3-task.tex
We present the \textbf{argument cloze} task, which allows us to automatically generate large scale data for training (Section \ref{sec::exp::implementation}) and evaluation (Section \ref{sec::datasets::cloze}).

In this task, we randomly remove an entity from an argument position of one event in the text. The entity in question needs to appear in at least one other place in the text. The task is then for the model to pick, from all entities appearing in the text, the one that has been removed. We first define what we mean by an event, then what we mean by an entity. Like \citet{Pichotta2016AAAI,Granroth2016AAAI}, we define an \textit{event} $e$ as consisting of a verbal predicate $v$, a subject $s$, a direct object $o$, and a prepositional object $p$ (along with the preposition). Here we only allow one prepositional argument in the structure, to avoid variable length input in the event composition model.\footnote{In case of multiple prepositional objects, we select the one that is closest to the predicate.} By an \textit{entity}, we mean a coreference chain with a length of at least two -- that is, the entity needs to appear at least twice in the text.

\begin{figure}[!ht]
\centering
\begin{subfigure}[b]{1\linewidth}
\includegraphics[width=\linewidth]{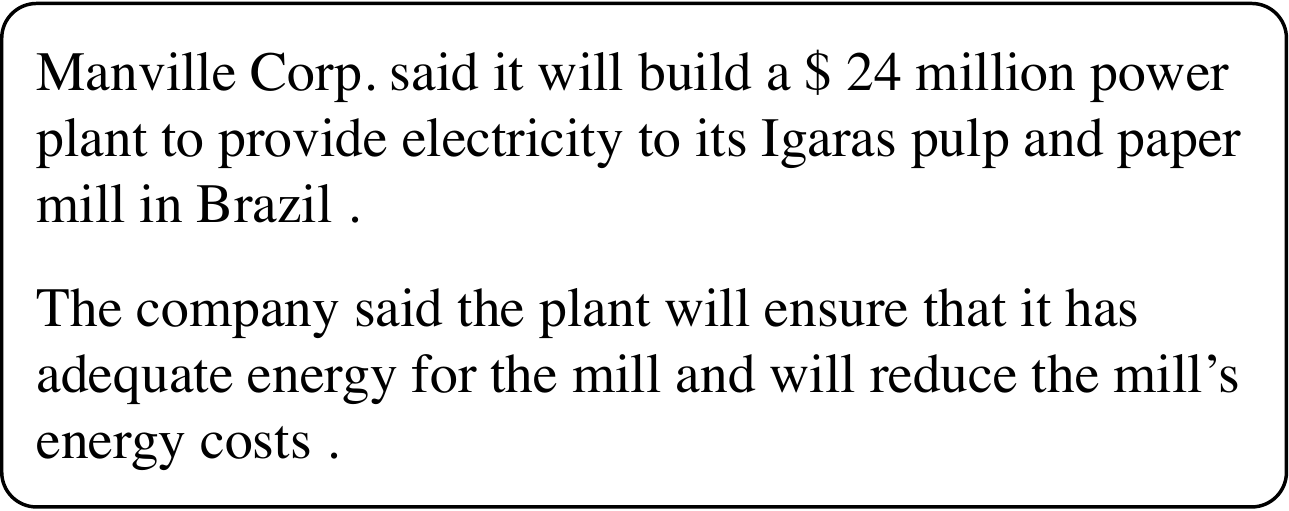}
\caption{A piece of raw text from OntoNotes corpus.}
\label{fig::script-example-a}
\vspace{6pt}
\end{subfigure}
~
\begin{subfigure}[b]{1\linewidth}
\includegraphics[width=\linewidth]{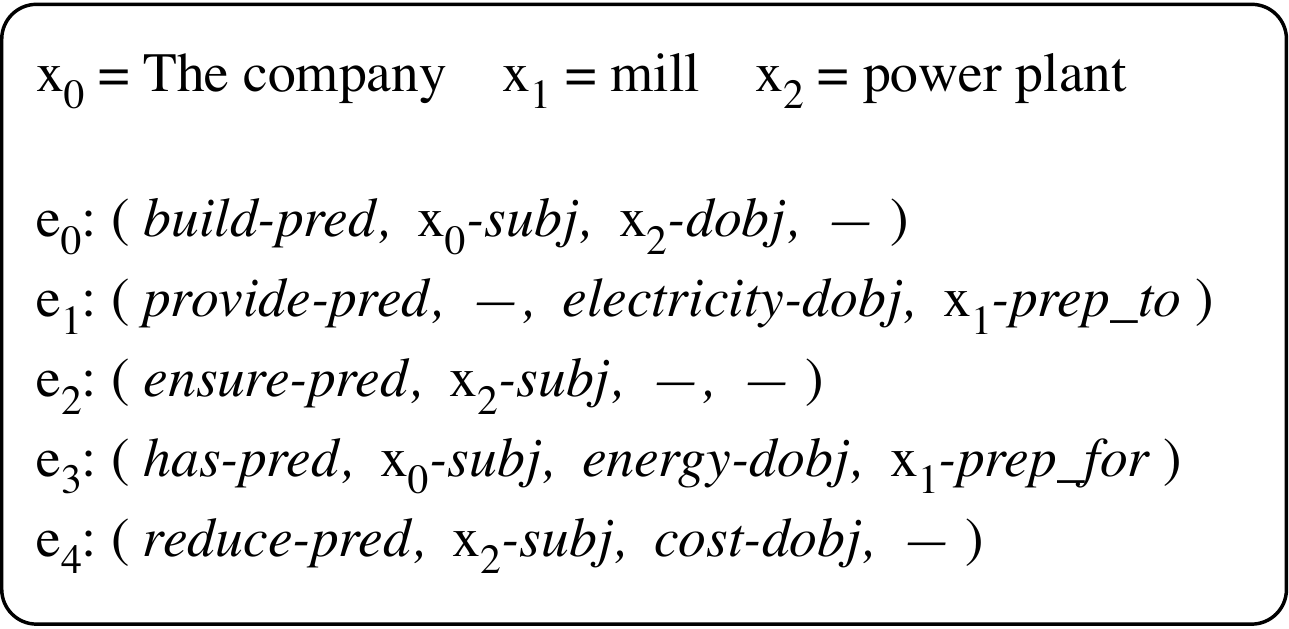}
\caption{Extracted events ($e_0$\texttildelow $e_4$) and entities ($x_0$\texttildelow $x_2$), using gold annotations from OntoNotes.}
\label{fig::script-example-b}
\vspace{6pt}
\end{subfigure}
~
\begin{subfigure}[b]{1\linewidth}
\includegraphics[width=\linewidth]{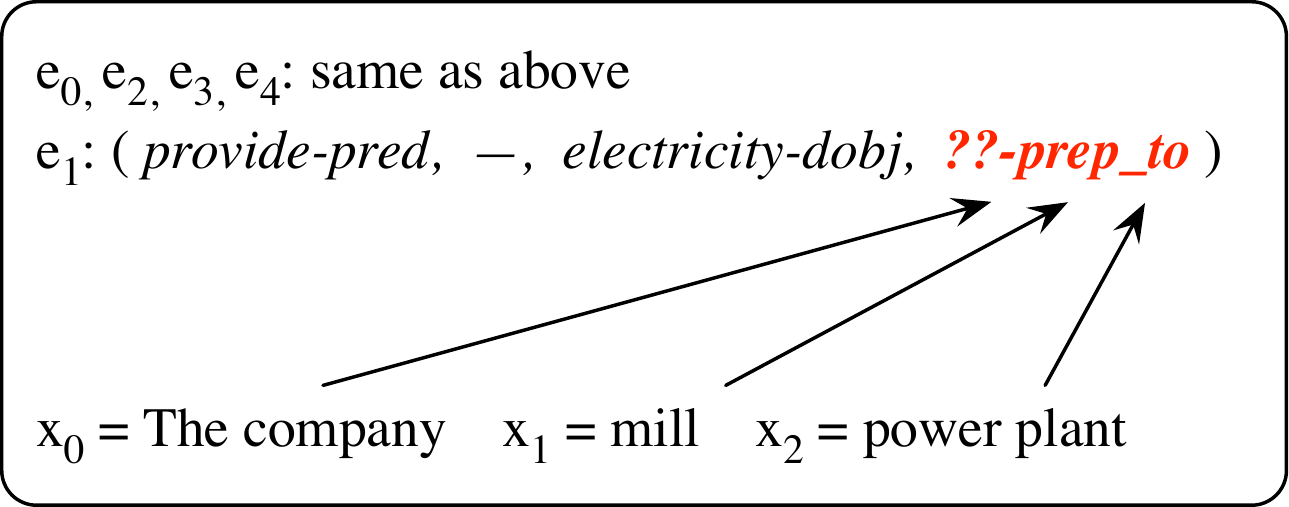}
\caption{Example of an argument cloze task for \emph{prep\_to} of $e_1$.}
\label{fig::script-example-c}
\end{subfigure}
\caption{Example of automatically extracted events and entities and an argument cloze task.}
\label{fig::script-example}
\end{figure}

For example, from a piece of raw text (Figure \ref{fig::script-example-a}), we automatically extract a sequence of events from a dependency parse, and a list of entities from coreference chains. In Figure \ref{fig::script-example-b}, $e_0$\texttildelow $e_4$ are events, $x_0$\texttildelow $x_2$ are entities. The arguments \textit{electricity-dobj} and \textit{energy-dobj} are not in coreference chains and are thus not candidates for removal. 
An example of the argument cloze task is shown in Figure \ref{fig::script-example-c}. Here the \emph{prep\_to} argument of $e_1$ has been removed. 

Coreference resolution is very noisy. Therefore we use gold coreference annotation for creating evaluation data, but automatically generated coreference chains for creating training data.

%% file: 4-methods.tex
\subsection{Modeling Narrative Coherence}
\label{sec::methods::predict}

We model implicit argument prediction as selecting the entity that, when filled in as the implicit argument, makes the overall most coherent narrative. Suppose we are trying to predict the direct object argument of some target event $e_t$. Then we complete $e_t$ by putting an entity candidate into the direct object argument position, and check the coherence of the resulting event with the rest of the narrative. Say we have a sequence of events $e_1, e_2, \dots, e_n$ in a narrative, and a list of entity candidates $x_1, x_2, \dots, x_m$. Then for any candidate $x_j$, we first complete the target event to be
\begin{equation}
\label{eq::target-event}
e_t(j) = (v_t, s_t, x_j, p_t),\ \ j = 1, \dots, m
\end{equation}
where $v_t$, $s_t$, and $p_t$ are the predicate, subject, and prepositional object of $e_t$ respectively, and $x_j$ is filled as the direct object. (Event completion for omitted subjects and prepositional objects is analogous.)

Then we compute the narrative coherence score $S_j$ of the candidate $x_j$ by\footnote{We have also tried using the sum instead of the maximum, but it did not perform as well across different models and datasets.}
\begin{equation}
S_j = \max_{c=1,\ c\neq t}^n coh\left(\vec{e_t(j)}, \vec{e_c}\right),\ \ j = 1, \dots, m
\end{equation}
where $\vec{e_t(j)}$ and $\vec{e_c}$ are representations for the completed target event $e_t(j)$ and one context event $e_c$, and $coh$ is a function computing a coherence score between two events, both depending on the model being used. The candidate $x_j$ with the highest score $S_j$ is then selected as our prediction.





\subsection{The Event Composition Model}
\label{sec::methods::event-comp}
\begin{figure*}[!ht]
\centering
\includegraphics[width=\linewidth]{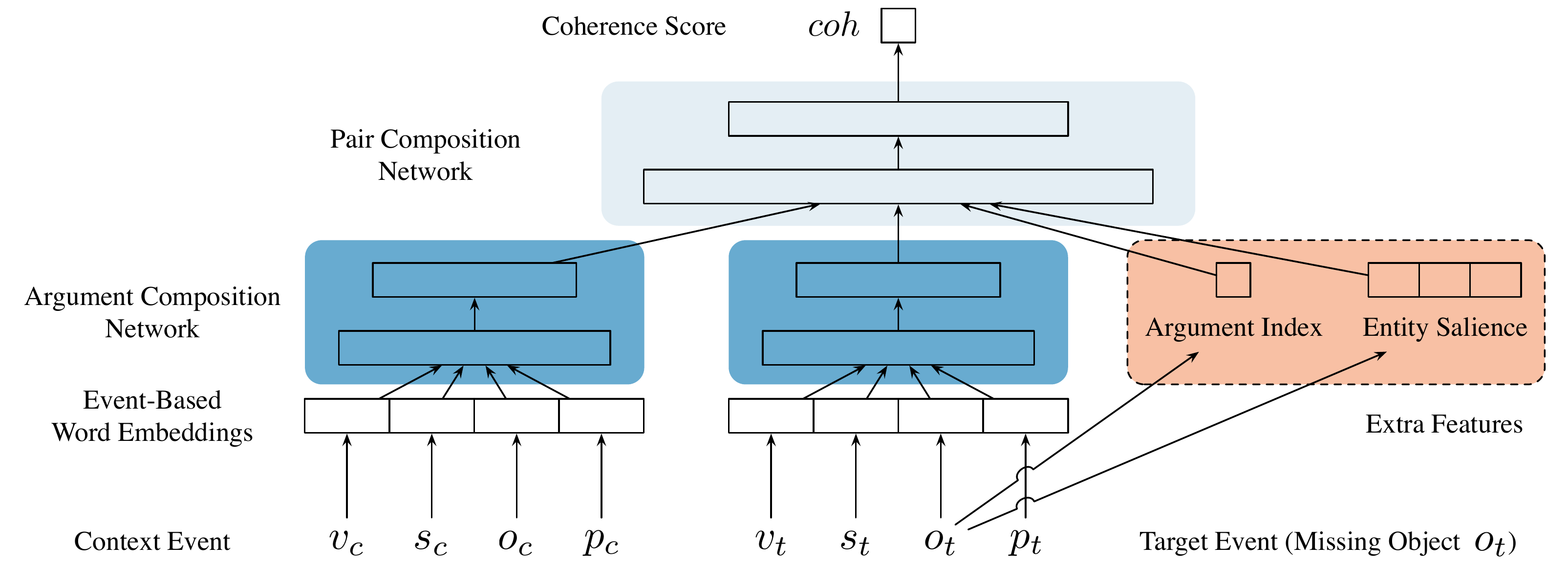}
\caption{Diagram for event composition model. \textbf{\emph{Input}}: a context event and a target event. \textbf{\emph{Event-Based Word Embeddings}}: embeddings for components of both events that encodes event knowledge. \textbf{\emph{Argument Composition Network}}: produces an event representation from its components. \textbf{\emph{Pair Composition Network}}: computes a coherence score $coh$ from two event representations. \textbf{\emph{Extra Features}}: argument index and entity salience features as additional input to the pair composition network.}
\label{fig::event-comp-model}
\end{figure*}

To model coherence ($coh$) between a context event and a target event,
we build an event composition model consisting of three parts, as shown in Figure \ref{fig::event-comp-model}: event components are representated through \textbf{event-based word embeddings}, which encode event knowledge in word representations; the \textbf{argument composition network} combines the components to produce event representations; and the \textbf{pair composition network} compute a coherence score for two event representations.

This basic architecture is as in the model of \citet{Granroth2016AAAI}. However our model is designed for a different task, argument cloze rather than narrative cloze, and for our task entity-specific information is more important. We therefore create the training data in a different way, as described in Section \ref{sec::methods::training}. We now discuss the three parts of the model in more detail. 

\begin{figure}[!htb]
\centering
\includegraphics[width=\linewidth]{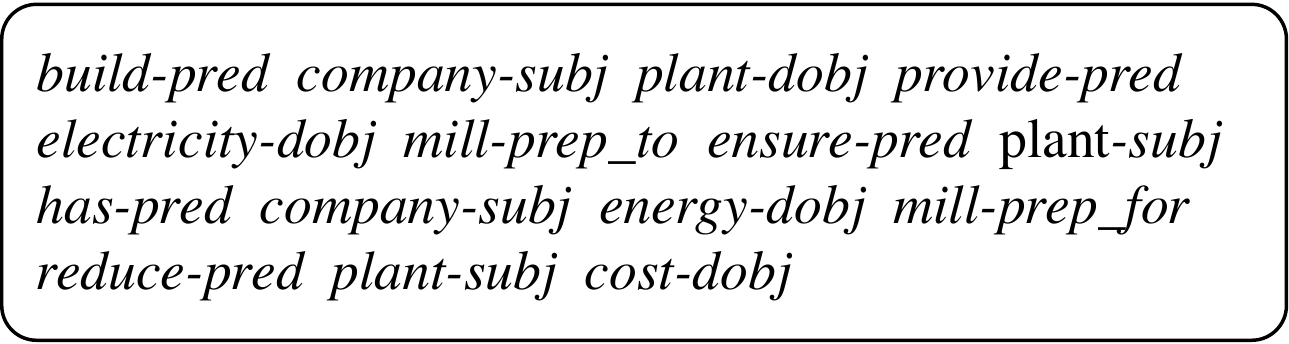}
\caption{Event-based word2vec training sentence, constructed from events and entities in Figure \ref{fig::script-example-b}.}
\label{fig::word2vec}
\end{figure}

\paragraph{Event-Based Word Embeddings}
The model takes word embeddings of both predicates and arguments as input to compute event representations. To better encode event knowledge in word level, we train an SGNS (skip-gram with negative sampling) word2vec model \cite{mikolov2013word2vec} with event-specific information. For each extracted event sequence, we create a sentence with the predicates and arguments of all events in the sequence.  An example of such a training sentence is given in Figure \ref{fig::word2vec}.


\paragraph{Argument Composition Network}
The argument composition network (dark blue area in Figure \ref{fig::event-comp-model}) is a two-layer feedforward neural network that composes an event representation from the embeddings of its components. Non-existent argument positions are filled with zeros.

\paragraph{Pair Composition Network}
The pair composition network (light blue area in Figure \ref{fig::event-comp-model}) computes a coherence score $coh$ between 0 and 1, given the vector representations of a context event and a target event. The coherence score should be high when the target event contains the correct argument, and low otherwise. So we construct the training objective function to distinguish the correct argument from wrong ones, as described in Equation \ref{eq::objective}.

\subsubsection{Training for Argument Prediction}
\label{sec::methods::training}
\begin{figure}[!htb]
\centering
\includegraphics[width=\linewidth]{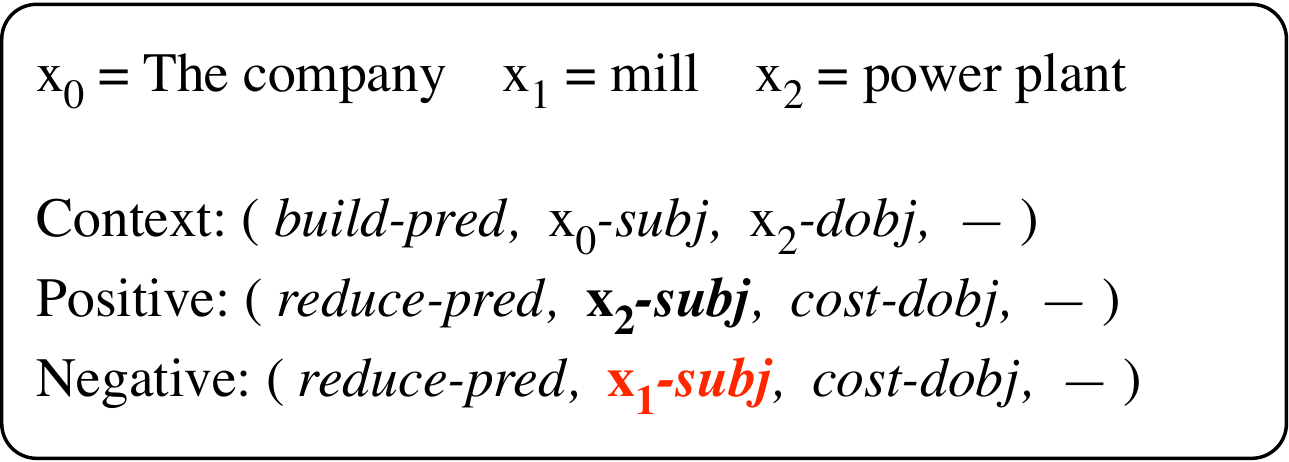}
\caption{Example of an event triple constructed from events and entities in Figure \ref{fig::script-example-b}.}
\label{fig::event-triple}
\end{figure}

To train the model to pick the correct candidate, we automatically construct training samples as event triples consisting of a context event $e_c$, a positive event $e_p$, and a negative event $e_n$. The context event and positive event are randomly sampled from an observed sequence of events, while the negative event is generated by replacing one argument of positive event by a random entity in the narrative, as shown in Figure \ref{fig::event-triple}.

We want the coherence score between $e_c$ and $e_p$ to be close to $1$, while the score for $e_c$ and $e_n$ should be close to $0$. Therefore, we train the model to minimize cross-entropy as follows:
\begin{equation}
\label{eq::objective}
\dfrac{1}{m}\sum_{i=1}^m -\log(coh(e_{ci}, e_{pi})) - \log(1 - coh(e_{ci}, e_{ni}))
\end{equation}
where $e_{ci}$, $e_{pi}$, and $e_{ni}$ are the context, positive, and negative events of the $i$th training sample respectively.

\subsection{Entity Salience}
\label{sec::methods:salience}
Implicit arguments tend to be salient entities in the document. So we extend our model by entity salience features, building on recent work by \citet{Dunietz2014EACL}, who introduced a simple model with several surface level features for entity salience detection. 
Among the features they used, we discard those that require external resources, and only use the remaining three features, as illustrated in Table \ref{tab::entity-salience}. Dunietz and Gillick found \emph{mentions} to be the most powerful indicator for entity salience among all features. We expect similar results in our experiments, however we include all three features in our event composition model for now, and conduct an ablation test afterwards.

\begin{table}[!ht]
\centering
\begin{tabularx}{\linewidth}{l X}
\hline
Feature & Description \\
\hline
\emph{1st\_loc} & Index of the sentence where the first mention of the entity appears \\
\emph{head\_count} & Number of times the head word of the entity appears \\
\emph{mentions} & A vector containing the numbers of named, nominal, pronominal, and total mentions of the entity \\
\hline
\end{tabularx}
\caption{Entity salience features from \citet{Dunietz2014EACL}.}
\label{tab::entity-salience}
\end{table}

The entity salience features are directly passed into the pair composition network as additional input. We also add an extra feature for argument position index (encoding whether the missing argument is a subject, direct object, or prepositional object), as shown in the red area in Figure \ref{fig::event-comp-model}.

%% file: 5-datasets.tex
\subsection{Argument Cloze Evaluation}
\label{sec::datasets::cloze}

Previous implicit argument datasets were very small. To overcome that limitation, we automatically create a large and comprehensive evaluation dataset, following the \textbf{argument cloze} task setting in Section \ref{sec::task}.

Since the events and entities are extracted from dependency labels and coreference chains, we do not want to introduce systematic error into the evaluation from imperfect parsing and coreference algorithms. Therefore, we create the evaluation set from OntoNotes \cite{Hovy2006Ontonotes}, which contains human-labeled dependency and coreference annotation for a large corpus. So the extracted events and entities in the evaluation set are gold. Note that this is only for evaluation; in training we do not rely on any gold annotations (Section \ref{sec::exp::implementation}).

There are four English sub-corpora in OntoNotes Release 5.0\footnote{LDC Catalog No. LDC2013T19} that are annotated with dependency labels and coreference chains. Three of them, which are mainly from broadcast news, share similar statistics in document length, so we combine them into a single dataset and name it \textbf{\textsc{ON-Short}} as it consists mostly of short documents. The fourth subcorpus is from the \emph{Wall Street Journal} and has significantly longer documents. We call this subcorpus \textbf{\textsc{ON-Long}} and evaluate on it separately. Some statistics are shown in Table \ref{tab::synthetic-data}.

\begin{table}[!htb]
\centering
\begin{tabularx}{\linewidth}{c X X}
\hline
& \textsc{ON-Short} & \textsc{ON-Long} \\
\hline
\# doc & 1027 & 597 \\
\# test cases & 13018 & 18208 \\
Avg \# entities & 12.06 & 36.95 \\
\hline
\end{tabularx}
\caption{Statistics on argument cloze datasets.}
\label{tab::synthetic-data}
\end{table}

\subsection{The Gerber and Chai (G\&C) Dataset}
\label{sec::datasets::gerber-chai}

The implicit argument dataset from \citet{Gerber2010ACL} (referred as \textbf{\textsc{G\&C}} henceforth) consists of 966 human-annotated implicit argument instances on 10 nominal predicates.

To evaluate our model on \textsc{G\&C}, we convert the annotations to the input format of our model as follows:
We map nominal predicates to their verbal form, and semantic role
labels to syntactic argument types based on the NomBank frame
definitions. One of the examples (after mapping semantic role labels) is as follows:

\begin{quote}
[Participants]\textsubscript{\emph{subj}} will be able to transfer [money]\textsubscript{\emph{dobj}} to [other investment funds]\textsubscript{\emph{prep\_to}}. The [investment]\textsubscript{\emph{pred}} choices are limited to [a stock fund and a money-market fund]\textsubscript{\emph{prep\_to}}.
\end{quote}

For the nominal predicate \emph{investment}, there are three arguments missing (\emph{subj}, \emph{dobj}, \emph{prep\_to}). The model first needs to determine that each of those argument positions in fact has an implicit filler. Then, from a list of candidates (not shown here), it needs to select \emph{Participants} as the implicit \emph{subj} argument, \emph{money} as the implicit \emph{dobj} argument, and either \emph{other investment funds} or \emph{a stock fund and a money-market fund} as the implicit \emph{prep\_to}.



%

%% file: 6-experiments.tex
\subsection{Implementation Details}
\label{sec::exp::implementation}

We train our neural model using synthetic data as described in Section~\ref{sec::task}. For creating the training data, we do not use gold parses or gold coreference chains.
We use the 20160901 dump of English Wikipedia\footnote{\url{https://dumps.wikimedia.org/enwiki/}}, with 5,228,621 documents in total. For each document, we extract plain text and break it into paragraphs, while discarding all structured data like lists and tables\footnote{We use the WikiExtractor tool at \url{https://github.com/attardi/wikiextractor}.}.
We construct a sequence of events and entities from each paragraph, by running Stanford CoreNLP~\cite{Manning2014ACL} to obtain dependency parses and coreference chains. We lemmatize all verbs and arguments. We incorporate negation and particles in verbs, and normalize passive constructions. We represent all arguments by their entity indices if they exist, otherwise by their head lemmas. We keep verbs and arguments with counts over 500, together with the 50 most frequent prepositions, leading to a vocabulary of 53,345 tokens; all other words are replaced with an out-of-vocabulary token. The most frequent verbs (with counts over 100,000) are down-sampled.

For training the event-based word embeddings, we create pseudo-sentences (Section \ref{sec::methods::event-comp}) from all events of all sequences (approximately 87 million events) as training samples.
We train an SGNS word2vec model with embedding size = $300$, window size = $10$, subsampling threshold = $10^{-4}$, and negative samples = $10$, using the Gensim package \cite{rehurek_lrec}.

For training the event composition model, we follow the procedure described in Section \ref{sec::methods::training}, and extract approximately 40 million event triples as training samples\footnote{We only sample one negative event for each pair of context and positive events for fast training, though more training samples are easily accessible.}. We use a two-layer feedforward neural network with layer sizes 600 and 300 for the argument composition network, and another two-layer network with layer sizes 400 and 200 for the pair composition network. We use cross-entropy loss with $\ell_2$ regularization of 0.01. We train the model using stochastic gradient descent (SGD) with a learning rate of 0.01 and a batch size of 100 for 20 epochs.

To study how the size of the training set affects performance, we
downsample the 40 million training samples to another set of 8 million
training samples. We refer to the resulting models as \textbf{\textsc{EventComp-8M}} and 
\textbf{\textsc{EventComp-40M}}.





\subsection{Evaluation on Argument Cloze}
\label{sec::exp::cloze}

For the synthetic argument cloze task, we compare our model with 3 baselines.

\vspace{-3pt}
\paragraph{\textsc{Random}} Randomly select one entity from the candidate list.
\vspace{-3pt}
\paragraph{\textsc{MostFreq}} Always select the entity with highest number of mentions.
\vspace{-3pt}
\paragraph{\textsc{EventWord2vec}} Use the event-based word embeddings
described in Section \ref{sec::methods::event-comp} for predicates and
arguments. The representation of an event $e$ is the sum of the embeddings
of its components, i.e.,
\begin{equation}
\vec{e} = \vec{v} + \vec{s} + \vec{o} + \vec{p}
\end{equation}
where $\vec{v}, \vec{s}, \vec{o}, \vec{p}$ are the embeddings of verb, subject, object, and
prepositional object, respectively. The coherence score of two events
in this baseline model is their cosine similarity. 
Like in our main model, the coherence score of the candidate is then the
maximum pairwise coherence score, as described in Section \ref{sec::methods::predict}.

The evaluation results on the \textsc{ON-Short} dataset are shown in
Table \ref{tab::eval-on-short}. The \textsc{EventWord2vec} baseline is
much stronger than the other two, achieving an accuracy of 38.40\%.
In fact, \textsc{EventComp-8M} by itself does not do better
than \textsc{EventWord2vec}, but adding entity salience greatly boosts
performance. Using more training data (\textsc{EventComp-40M}) helps
by a substantial margin both with and without entity salience
features. 

\begin{table}[!htb]
\centering
\setlength{\tabcolsep}{12pt}
\begin{tabular}{l l}
\hline
\noalign{\vskip 1.5pt}
& Accuracy (\%) \\
\noalign{\vskip 1.5pt}
\hline
\noalign{\vskip 1.5pt}
\textsc{Random} & 8.29 \\
\textsc{MostFreq} & 22.76 \\
\textsc{EventWord2vec} & 38.40 \\
\noalign{\vskip 1.5pt}
\hline
\noalign{\vskip 1.5pt}
\textsc{EventComp-8M} & 38.26 \\
\quad + entity salience & 45.05 \\
\textsc{EventComp-40M} & 41.89 \\
\quad + entity salience & \textbf{47.75} \\
\noalign{\vskip 1.5pt}
\hline
\end{tabular}
\caption{Evaluation on \textsc{ON-Short}.}
\vspace{-3pt}
\label{tab::eval-on-short}
\end{table}

To see which of the entity salience features are important, we conduct
an ablation test with the \textsc{EventComp-8M} model on
\textsc{ON-Short}. From the results in Table \ref{tab::ablation}, we
can see that in our task, as in \citet{Dunietz2014EACL}, the entity
mentions features,  i.e., the numbers of named, nominal, pronominal,
and total mentions of the entity,  are most helpful. In fact, the
other two features even decrease performance slightly. 

\begin{table}[ht]
\centering
\setlength{\tabcolsep}{12pt}
\begin{tabular}{l l}
\hline
\noalign{\vskip 1.5pt}
Features & Accuracy (\%) \\
\noalign{\vskip 1.5pt}
\hline
\noalign{\vskip 1.5pt}
no entity salience feature & 38.26 \\
\noalign{\vskip 1.5pt}
\hline
\noalign{\vskip 1.5pt}
\quad -- \textit{mentions} & 39.02 \\
\quad -- \textit{head\_count} & \textbf{45.71} \\
\quad -- \textit{1st\_loc} & \textbf{45.65} \\
\noalign{\vskip 1.5pt}
\hline
\noalign{\vskip 1.5pt}
all entity salience features & 45.05 \\
\noalign{\vskip 1.5pt}
\hline
\end{tabular}
\caption{Ablation test on entity salience features. (Using \textsc{EventComp-8M} on \textsc{ON-Short}.)}
\vspace{-3pt}
\label{tab::ablation}
\end{table}

\begin{figure*}[!htb]
	\centering
	\begin{subfigure}[b]{0.32\textwidth}
		\includegraphics[width=\linewidth]{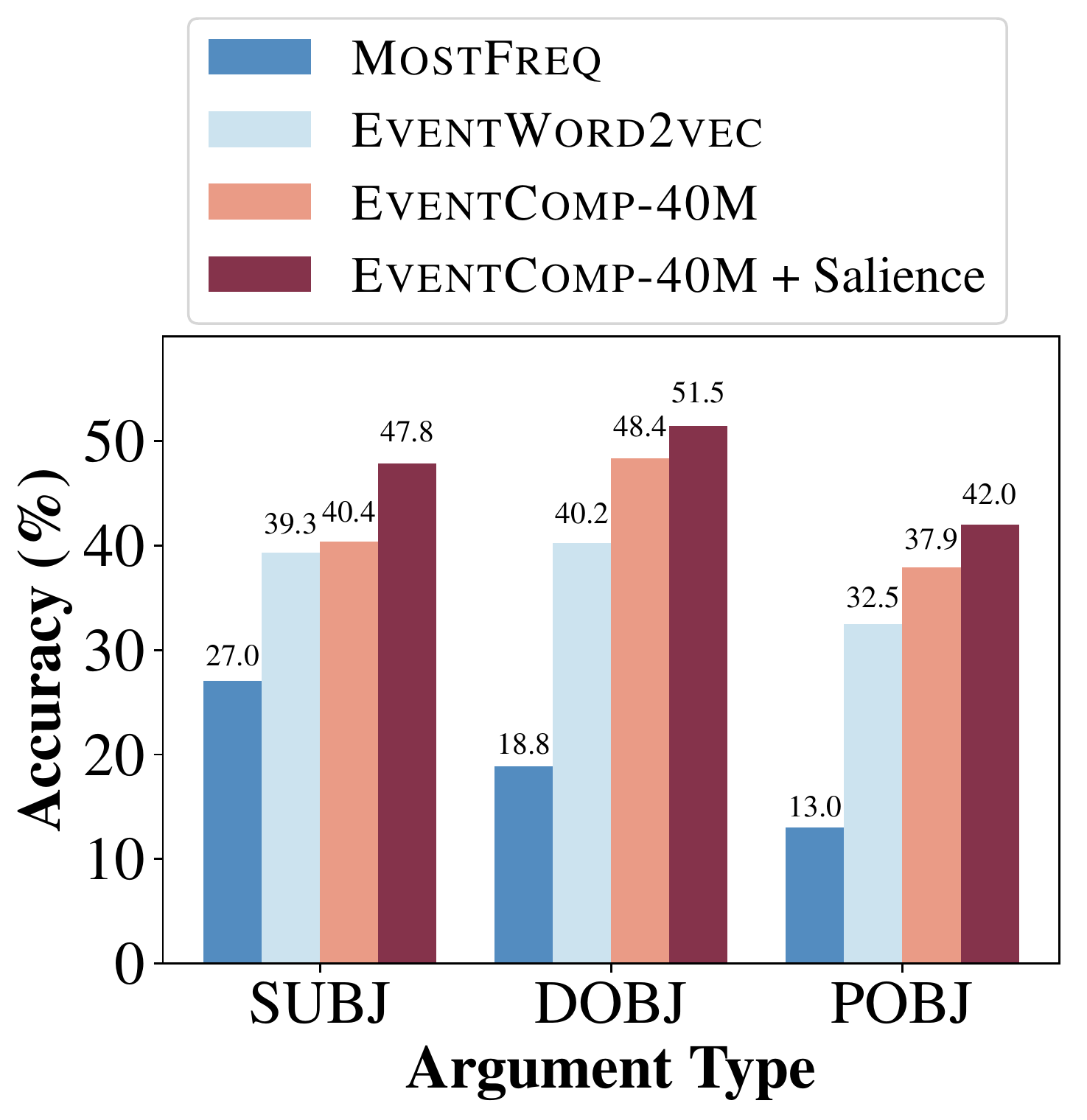}
		\caption{Accuracy by Argument Type}
		\label{fig::case-analysis::type}
	\end{subfigure}
	~
	\begin{subfigure}[b]{0.32\textwidth}
		\includegraphics[width=\linewidth]{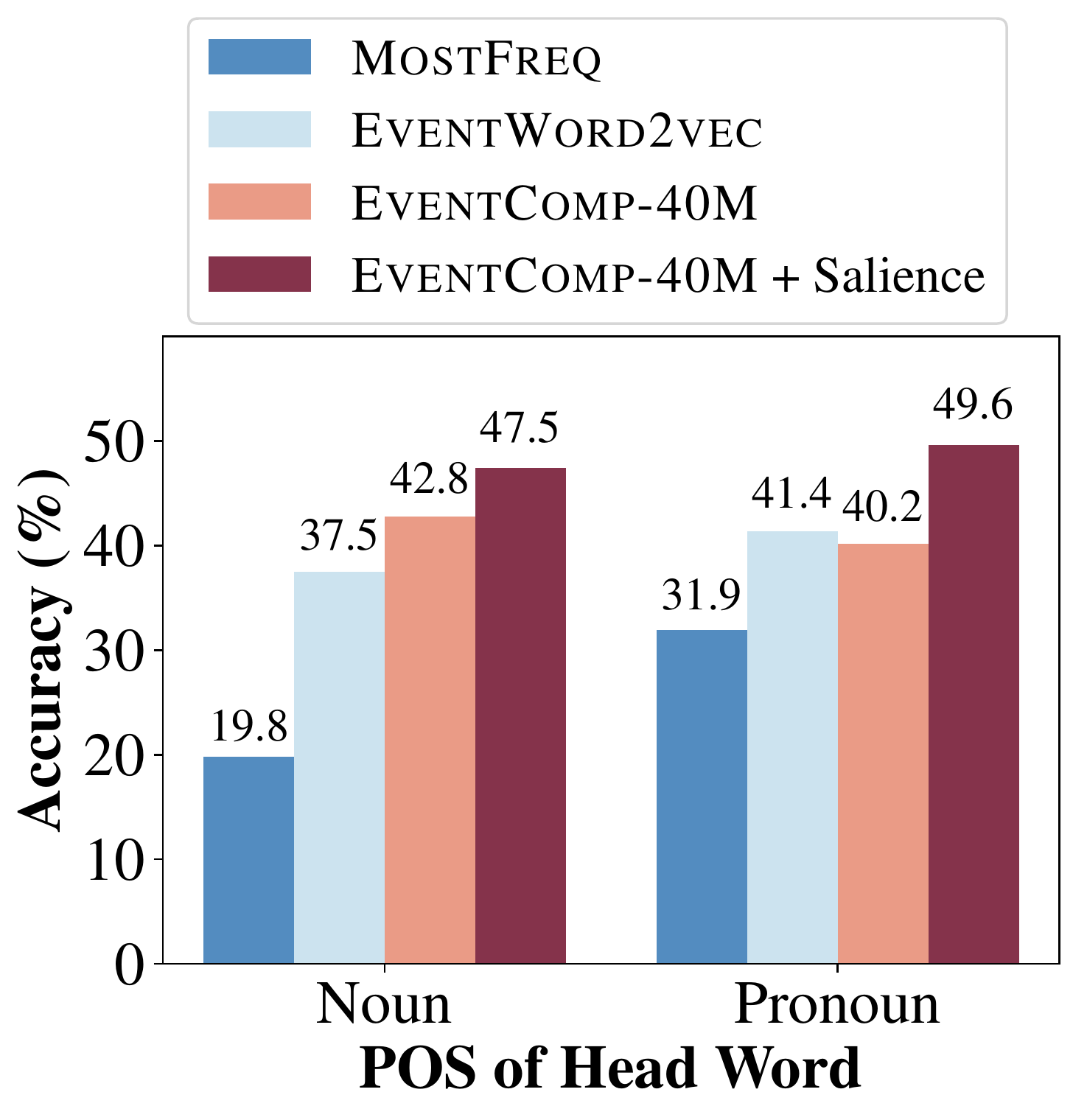}
		\caption{Accuracy by POS of Head Word}
		\label{fig::case-analysis::pos}
	\end{subfigure}
	~
	\begin{subfigure}[b]{0.32\textwidth}
		\includegraphics[width=\linewidth]{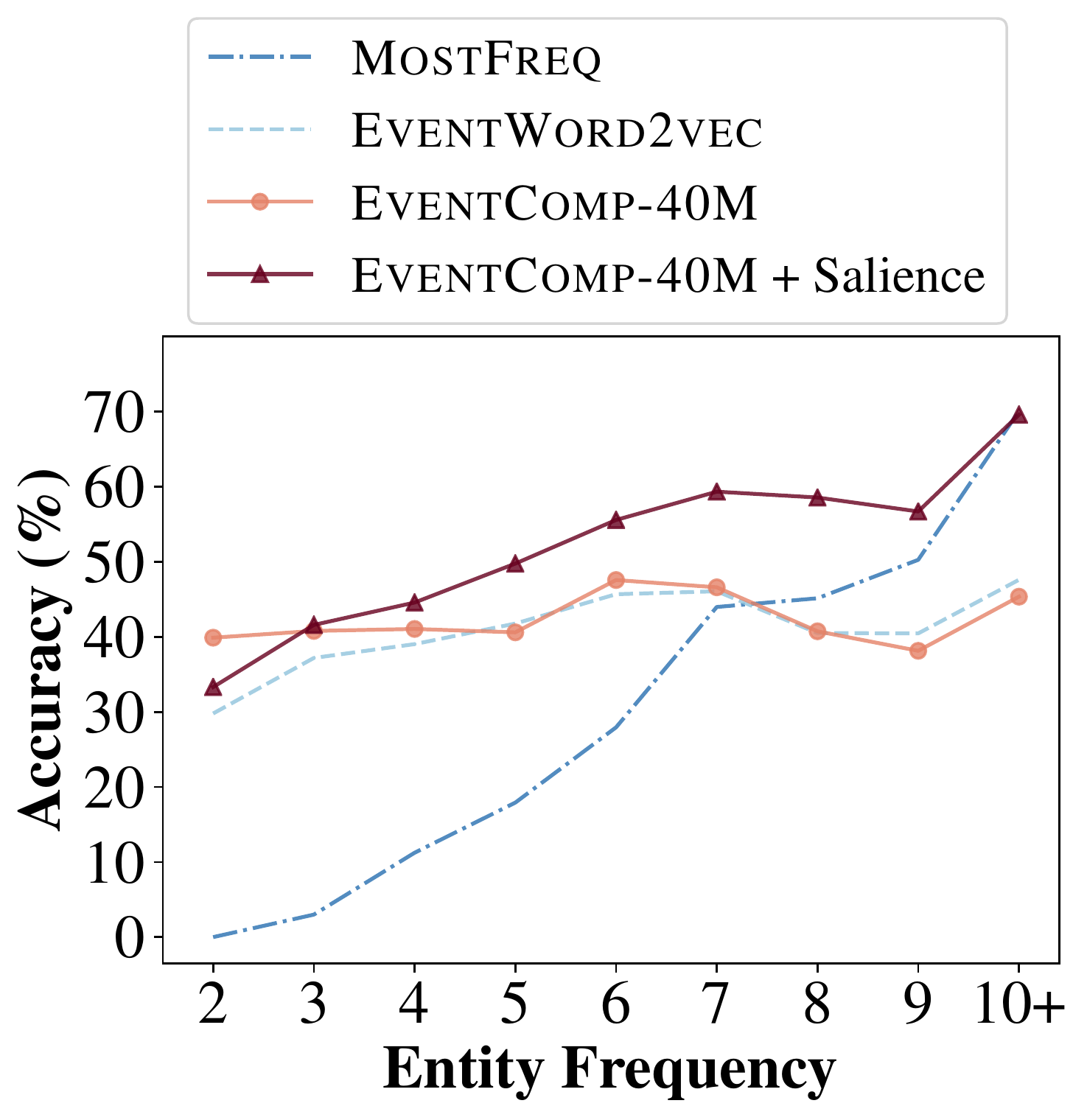}
		\caption{Accuracy by Entity Frequency}
		\label{fig::case-analysis::frequency}
	\end{subfigure}
	\caption{Performance of \textsc{EventComp} (with and without entity salience) and two baseline models by (a) argument type, (b) part-of-speech tag of the head word of the entity, and (c) entity frequency.}
	\label{fig::case-analysis}
\end{figure*}

We take a closer look at several of the models in Figure~\ref{fig::case-analysis}. Figure~\ref{fig::case-analysis::type}
breaks down the results by the argument type of the removed
argument. On subjects, the \textsc{EventWord2vec}
baseline matches the performance of \textsc{EventComp}, but not on direct objects and prepositional
objects. Subjects are semantically much less diverse than the other
argument types, as they are very often animate. A similar pattern
is apparent in Figure~\ref{fig::case-analysis::pos}, which has results by the
part-of-speech tag of the head word of the removed entity. Note that an entity is
a coreference chain, not a single mention; so when the head word is a
pronoun, this is an entity which has only pronoun mentions. A pronoun
entity provides little semantic content beyond, again, animacy. And
again, \textsc{EventWord2vec} performs well on pronoun entities, but
less so on entities described by a noun. It seems that
\textsc{EventWord2vec} can pick up on a coarse-grained pattern such as
animate/inanimate, but not on more fine-grained distinctions needed to
select the right noun, or to select a fitting direct object or
prepositional object. This matches the fact that
\textsc{EventWord2vec} gets a less clear signal on the task, in two
respects: It gets much less information than \textsc{eventComp} on the distinction between
argument positions,\footnote{As shown in Figure~\ref{fig::word2vec},
  the ``words'' for which embeddings are computed are role-lemma pairs.}  and it only looks at overall event
similarity while \textsc{EventComp} is trained to detect narrative
coherence. Entity salience contributes greatly across all
argument types and parts of speech, but more strongly on subjects and
pronouns. This is again because subjects, and pronouns,
are semantically less distinct, so they can only be distinguished by
relative salience.





Figure \ref{fig::case-analysis::frequency} analyzes results by the
frequency of the removed entity, that is, by its number of mentions. The
\textsc{MostFreq} baseline, unsurprisingly, only does well when the
removed entity is a highly frequent one. The \textsc{EventComp}
model is much better than \textsc{MostFreq} at picking out the right
entity when it is a rare one, as it can look at the semantic content
of the entity as well as its frequency. Entity salience boosts the
performance of \textsc{EventComp} in particular for frequent
entities. 



The \textsc{ON-Long} dataset, as discussed in
Section~\ref{sec::datasets::cloze}, consists of OntoNotes data with
much longer documents than found in \textsc{ON-Short}. Evaluation
results on \textsc{ON-Long} are shown in Table
\ref{tab::eval-on-long}. Although the overall numbers are lower than those for 
\textsc{ON-Short}, we are selecting from $36.95$ candidates on
average, more than 3 times more than for 
\textsc{ON-Short}. Considering that the accuracy of randomly selecting
an entity is as low as $2.71\%$, the performance of our best
performing model, with an
accuracy of $27.87\%$, is quite good. 

\begin{table}[!htb]
\centering
\setlength{\tabcolsep}{12pt}
\begin{tabular}{l l}
\hline
\noalign{\vskip 1.5pt}
& Accuracy (\%) \\
\noalign{\vskip 1.5pt}
\hline
\noalign{\vskip 1.5pt}
\textsc{Random} & 2.71 \\
\textsc{MostFreq} & 17.23 \\
\textsc{EventWord2vec} & 21.49 \\
\noalign{\vskip 1.5pt}
\hline
\noalign{\vskip 1.5pt}
\textsc{EventComp-8M} & 18.79 \\
\quad + entity salience & 26.23\\
\textsc{EventComp-40M} & 21.79 \\
\quad + entity salience & \textbf{27.87} \\
\noalign{\vskip 1.5pt}
\hline
\end{tabular}
\caption{Evaluation on \textsc{ON-Long}.}
\label{tab::eval-on-long}
\end{table}

\subsection{Evaluation on \textsc{G\&C}}
\label{sec::exp::gerber-chai}

The G\&C data differs from the Argument Cloze data in two
respects. First, not every argument position that seems to be open
needs to be filled: The model must additionally make a
\textbf{fill / no-fill decision}. Whether a particular argument position
is typically filled is highly predicate-specific. As the small G\&C
dataset does not provide enough data to train our neural model on this
task, we instead train a simple logistic classifier, the 
\textbf{fill / no-fill classifier}, with a small subset of shallow lexical features
used in \citet{Gerber2012CL}, to make the decision. These features
describe the syntactic context of the predicate. We use only 14 features;
the original Gerber and Chai model had more than 80 features, and our
re-implementation, described below, has around 60. 

\begin{figure}[!ht]
\centering
\includegraphics[width=\linewidth]{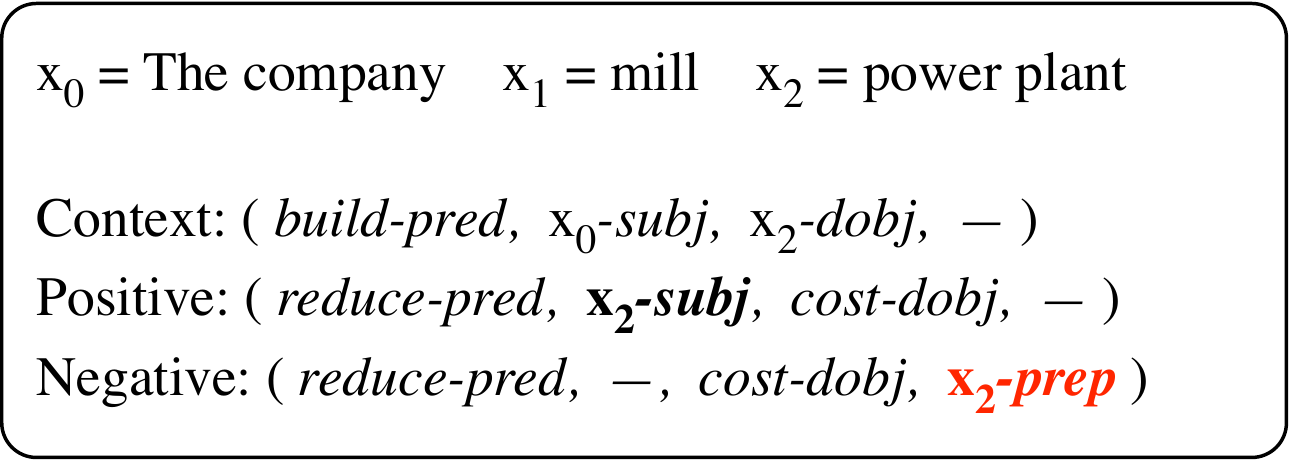}
\caption{Event triples for training multi implicit argument prediction.}
\label{fig::extra-training}
\end{figure}

The second difference is that in G\&C, an event may have multiple open argument
positions. In that case, the task is not just to select a candidate
entity, but also to determine which of the open argument positions it
should fill. So the model must do \textbf{multi implicit argument
  prediction}. We can flexibly adapt our method for training data
generation to this case. In particular, we create extra negative
training events, in which an argument of the positive event has been moved to
another argument position in the same event, as shown in Figure
\ref{fig::extra-training}. We can then simply train our
\textsc{EventComp} model on this extended training data. We refer to the extra training process as
\textbf{multi-arg training}. 

We compare our models to that of \citet{Gerber2012CL}. However, their original logistic regression model used
many features based on gold annotation from FrameNet, PropBank and
NomBank. To create a more realistic evaluation setup, we re-implement a variant of their original model by removing gold features, and name it \textsc{GCauto}. Results from \textsc{GCauto} are directly comparable to our models, as both are trained on automatically generated features.
\footnote{To be fair, we also tested adding the fill / no-fill classifier to \textsc{GCauto}. However the classifier only increases precision at the cost of reducing recall, and \textsc{GCauto} already has higher precision than recall. The resulting $F_1$ score is actually worse, and thus is not reported here.}


\begin{table}[!htb]
\centering
\begin{tabular}{l l l l}
\hline
\noalign{\vskip 1.5pt}
& $P$ & $R$ & $F_1$ \\
\noalign{\vskip 1.5pt}
\hline
\noalign{\vskip 1.5pt}
\citet{Gerber2012CL} & 57.9 & 44.5 & 50.3 \\
\quad \textsc{GCauto} & 49.9 & 40.1 & 44.5 \\
\noalign{\vskip 1.5pt}
\hline
\noalign{\vskip 1.5pt}
\textsc{EventComp-8M} & 8.9 & 27.9 & 13.5 \\
\quad + fill / no-fill classifier & 22.0 & 22.3 & 22.1 \\
\qquad + multi-arg training & 43.5 & 44.1 & 43.8 \\
\quad\qquad + entity salience & 45.7 & 46.4 & \textbf{46.1} \\
\noalign{\vskip 1.5pt}
\hline
\noalign{\vskip 1.5pt}
\textsc{EventComp-40M} & 9.4 & 30.3 & 14.3 \\
\quad + fill / no-fill classifier & 23.7 & 24.0 & 23.9 \\
\qquad +  multi-arg training & 46.7 & 47.3 & 47.0 \\
\quad\qquad + entity salience & 49.3 & 49.9 & \textbf{49.6} \\
\noalign{\vskip 1.5pt}
\hline
\end{tabular}
\caption{Evaluation on \textsc{G\&C} dataset.}
\label{tab::eval-gerber-chai}
\end{table}

We present the evaluation results in Table
\ref{tab::eval-gerber-chai}. The original \textsc{EventComp} models do
not perform well, which is as expected since the model is not designed
to do the \emph{fill / no-fill decision} and \emph{multi implicit 
argument prediction} tasks as described above. With the fill /
no-fill classifier, precision rises by around 13 points because this
classifier prevents many false positives. With additional multi-arg
training, $F_1$ score improves by another 22-23 points. At this point, our
model achieves a performance comparable to the much more complex
G\&C reimplementation \textsc{GCauto}. Adding entity
salience features further boosts both precision and recall, showing
that implicit arguments do tend to be filled by salient entities, as
we had hypothesized. Again, more training data substantially benefits
the task. Our best performing model, at 49.6 $F_1$, clearly
outperforms \textsc{GCauto}, and is comparable with the original
\citet{Gerber2012CL} model trained with gold features. \footnote{We also tried fine tune our model on the \textsc{G\&C} dataset with cross validation, but the model severely overfit, possibly due to the very small size of the dataset.}



%% file: 7-conclusion.tex
In this paper we have addressed the task of implicit argument
prediction. To support training at scale, we have introduced a simple
cloze task for which data can be generated automatically. 
We have introduced a neural model, which frames implicit argument
prediction as the task of selecting the textual entity that completes
the event in a maximally narratively coherent way. The model prefers
salient entities, where salience is mainly defined through the number of
mentions. Evaluating on synthetic data from OntoNotes, we find that
our model clearly outperforms even strong baselines, that salience is
important throughout for performance, and that event knowledge is
particularly useful for the (more verb-specific) object and
prepositional object arguments. Evaluating on the naturally occurring
data from Gerber and Chai, we find that in a comparison without gold
features, our model clearly outperforms the
previous state-of-the-art model, where again salience information is
important. 

The current paper takes a first step towards predicting implicit
arguments based on narrative coherence. We currently use a relatively simple model
for local narrative coherence; in the future we will turn to models that can
test global coherence for an implicit argument candidate. 
We also plan to investigate how the extracted implicit arguments can
be integrated into a downstream task that makes use of event
information, in particular we would like to experiment with reading
comprehension. 

%% file: supplemental.tex
\onecolumn
\section{Supplemental Material}
\label{sec::appendix}

\begin{table*}[!htb]
\centering
\begin{tabularx}{0.9\linewidth}{l  l  *{5}{X}}
	\hline
	\noalign{\vskip 1.5pt}
	Nominal Predicate & Verbal Form & $arg_0$ & $arg_1$ & $arg_2$ & $arg_3$ & $arg_4$ \\
	\noalign{\vskip 1.5pt}
	\hline
	\noalign{\vskip 1.5pt}
	bid & bid & \emph{subj} & \emph{prep\_for} & \emph{dobj} & -- & -- \\
	sale & sell & \emph{subj} & \emph{dobj} & \emph{prep\_to} & \emph{prep\_for} & \emph{prep} \\
	loan & loan & \emph{subj} & \emph{dobj} & \emph{prep\_to} & \emph{prep} & \emph{prep\_at} \\
	cost & cost & -- & \emph{subj} & \emph{dobj} & \emph{prep\_to} & \emph{prep} \\
	plan & plan & \emph{subj} & \emph{dobj} & \emph{prep\_for} & \emph{prep\_for} & -- \\
	investor & invest & \emph{subj} & \emph{dobj} & \emph{prep\_in} & -- & -- \\
	price & price & \emph{subj} & \emph{dobj} & \emph{prep\_at} & \emph{prep} & -- \\
	loss & lose & \emph{subj} & \emph{dobj} & \emph{prep\_to} & \emph{prep\_on} & -- \\
	investment & invest & \emph{subj} & \emph{dobj} & \emph{prep\_in} & -- & -- \\
	fund & fund & \emph{subj} & \emph{dobj} & \emph{prep} & \emph{prep\_on} & -- \\
	\noalign{\vskip 1.5pt}
	\hline
\end{tabularx}
\caption{Mappings from the 10 nominal predicates to their verbal forms, and mappings from the semantic role labels of each predicate to the corresponding dependency labels, as discussed in Section \ref{sec::datasets::gerber-chai}.}
\label{tab::mapping}
\end{table*}

\begin{table*}[!htb]
\centering
\begin{tabularx}{0.9\linewidth}{l X}
	\hline
	\noalign{\vskip 1.5pt}
	\# & Description \\
	\noalign{\vskip 1.5pt}
	\hline
	\noalign{\vskip 1.5pt}
	1 & $p$ itself. \\
	2 & $p$ \& $p$'s morphological suffix. \\
	3 & $p$ \& $iarg_n$. \\
	4 & Verbal form of $p$ \& $iarg_n$. \\
	5 & Frequency of $p$ within the document. \\
	6 & $p$ \& the stemmed content words in a one-word window around $p$. \\
	7 & $p$ \& the stemmed content words in a two-word window around $p$. \\
	8 & $p$ \& the stemmed content words in a three-word window around $p$. \\
	9 & $p$ \& whether $p$ is before a passive verb. \\
	10 & $p$ \& the head of the following prepositional phrase's object. \\
	11 & $p$ \& the syntactic parse tree path from $p$ to the nearest passive verb. \\
	12 & $p$ \& the part-of-speech of $p$'s parent's head word. \\
	13 & $p$ \& the last word of $p$'s right sibling. \\
	14 & Whether or not $p$'s left sibling is a quantifier (many, most, all, etc.). \\
	\noalign{\vskip 1.5pt}
	\hline
\end{tabularx}
\caption{Features used in the fill / no-fill classifier, as discussed in Section \ref{sec::exp::gerber-chai}. This is a subset of features used by \citet{Gerber2012CL}. Here, $p$ is the nominal predicate, $iarg_n$ is the integer $n$ of the semantic role label of the implicit argument, as shown in Table \ref{tab::mapping}, and the \& symbol denotes concatenation.} 
\label{tab::features}
\end{table*}